\documentclass[11pt]{article}

\usepackage{acl}

\usepackage{times}
\usepackage{latexsym}

\usepackage[T1]{fontenc}

\usepackage[utf8]{inputenc}

\usepackage{microtype}

\usepackage{bm}
\usepackage{mathtools}
\usepackage{multirow}
\usepackage{graphicx}
\usepackage{xhfill}
\usepackage{amssymb}
\usepackage{here}
\usepackage{pifont}
\usepackage[subrefformat=parens]{subcaption}
\usepackage{arydshln}
\usepackage{algorithm}
\usepackage{algpseudocode}
\usepackage{booktabs}

\newcommand{\figlabel}[1]{\label{fig:#1}}
\newcommand{\figref}[1]{Figure \ref{fig:#1}}
\newcommand{\tbllabel}[1]{\label{tbl:#1}}
\newcommand{\tblref}[1]{Table \ref{tbl:#1}}
\newcommand{\seclabel}[1]{\label{sec:#1}}
\newcommand{\secref}[1]{Section \ref{sec:#1}}
\newcommand{\equlabel}[1]{\label{eq:#1}}
\newcommand{\equref}[1]{Eq. (\ref{eq:#1})}

\newcommand{\mr}[2]{\multirow{#1}{*}{#2}}
\newcommand{\ml}[3]{\multicolumn{#1}{#2}{#3}}

\makeatletter
\newcommand{\algmargin}{\the\ALG@thistlm}
\makeatother
\newlength{\whilewidth}
\algnewcommand{\parRequire}[1]{\Require%
  \parbox[t]{\dimexpr\linewidth-\algmargin}{\strut #1\strut}}
\algnewcommand{\parState}[1]{\State%
  \parbox[t]{\dimexpr\linewidth-\algmargin}{\strut #1\strut}}

\title{Adaptive Natural Language Generation for Task-oriented Dialogue via Reinforcement Learning}

\author{
Atsumoto Ohashi \qquad Ryuichiro Higashinaka \\
Graduate School of Informatics, Nagoya University\\
\texttt{ohashi.atsumoto.c0@s.mail.nagoya-u.ac.jp} \\
\texttt{higashinaka@i.nagoya-u.ac.jp} \\
}

\begin{document}
\maketitle
\begin{abstract}
When a natural language generation (NLG) component is implemented in a real-world task-oriented dialogue system, it is necessary to generate not only natural utterances as learned on training data but also utterances adapted to the dialogue environment (e.g., noise from environmental sounds) and the user (e.g., users with low levels of understanding ability). Inspired by recent advances in reinforcement learning (RL) for language generation tasks, we propose ANTOR, a method for {\bf A}daptive {\bf N}atural language generation for {\bf T}ask-{\bf O}riented dialogue via {\bf R}einforcement learning. In ANTOR, a natural language understanding (NLU) module, which corresponds to the user's understanding of system utterances, is incorporated into the objective function of RL. If the NLG's intentions are correctly conveyed to the NLU, which understands a system's utterances, the NLG is given a positive reward. We conducted experiments on the MultiWOZ dataset, and we confirmed that ANTOR could generate adaptive utterances against speech recognition errors and the different vocabulary levels of users.
\end{abstract}

\section{Introduction}
In task-oriented dialogue systems, the role of the natural language generation (NLG) component is to convert a system's intentions, called dialogue acts (DAs), into natural language utterances and to convey DAs accurately to users \citep{10.1145/505282.505285, gao2019neural}. In recent years, data-driven language generation methods \citep{wen-etal-2015-semantically, peng2020few} using neural networks have been introduced to NLG for task-oriented dialogue systems, enabling natural utterance generation.

When such NLG is implemented in a realistic environment, however, it is essential to generate not only natural utterances as learned on training data but also utterances adapted to the dialogue environment and the user. For example, when interacting in a noisy environment, such as in a place with loud background noise or through a telephone, the system needs to use sentences and vocabulary that are less likely to be misrecognized. In addition, if the user is a child or a second language learner, it is necessary to generate utterances in plain terms that the user can easily understand. Therefore, it is essential for the NLG module to adaptively generate utterances for the dialogue environment and the user in real-world situations. However, it is challenging to implement optimal NLG using only supervised learning because it is not practical to create training data for every environment or user. Recently, for many generative tasks, such as machine translation, summary generation, and dialogue generation in open domains, many methods using reinforcement learning (RL) have been proposed. In these studies, non-differentiable objective functions, such as generated text quality and subjective user preferences, are used to optimize the language generation model and achieve high performance.

With this background in mind, this study proposes a method for {\bf A}daptive {\bf N}atural language generation for {\bf T}ask-{\bf O}riented dialogue via {\bf R}einforcement learning (ANTOR)\footnote{In this paper, ANTOR refers to both the method of fine-tuning NLG and the fine-tuned NLG model. Our code and data are publicly available at \url{https://github.com/nu-dialogue/antor}}, which adapts to the dialogue environment and the user. In our method, a reward function using a natural language understanding (NLU) model is set up, and a pre-trained NLG model is fine-tuned by using RL. That is, the NLG generates a system utterance for a given DA, and the NLU provides a positive reward if it can successfully recognize the original DA from the utterance. Experiments using the MultiWOZ dataset \citep{budzianowski-etal-2018-multiwoz} are conducted with multiple environments and users simulating real-world conditions, such as speech recognition errors and the different vocabulary levels of users. Our contribution is threefold:
\begin{itemize}
\item We propose ANTOR, a method for fine-tuning NLG for task-oriented dialogue via reinforcement learning. We conducted experiments using MultiWOZ to confirm that ANTOR can generate adaptive utterances for multiple NLUs with different model architectures.
\item Experiments were conducted in a noisy environment where speech recognition errors caused by background noise were simulated. The results show that ANTOR could generate utterances with words less likely to cause speech recognition errors.
\item Experiments were conducted using NLUs that simulated users with low vocabulary levels. The results confirmed that ANTOR was able to generate utterances using vocabulary appropriate for each vocabulary level.
\end{itemize}

\section{Related Work}
\subsection{Natural Language Generation for Task-oriented Dialogue}
Conventional NLG for task-oriented dialogues had used template-based and rule-based methods \citep{WALKER2002409, stent-etal-2004-trainable}. There, templates and rules had to be carefully designed manually by experts in each domain. Later, a data-driven method using machine learning was proposed \citep{OH2002387, angeli2010simple, mairesse-young-2014-stochastic}. \citet{kondadadi2013statistical} proposed a method for statistically generating utterances using k-means clustering and support vector machines. Recently, many generation models based on end-to-end learning have been proposed by using deep learning \citep{wen-etal-2016-multi, tran-nguyen-2017-natural, su-etal-2018-natural}. \citet{wen-etal-2015-semantically} proposed SC-LSTM, which controls utterance generation by using DA feature vectors and reading gates. SC-GPT \citep{peng2020few} is a state-of-the-art model for MultiWOZ that achieves high performance by fine-tuning the language model GPT-2 \citep{radford2019language} on a large number of task-oriented dialog datasets.

Some end-to-end models \citep{budzianowski-etal-2018-multiwoz, chen-etal-2019-semantically} generate system utterances directly from a dialogue history instead of using NLG, which is known as a word-level policy. In particular, \citet{zhao-etal-2019-rethinking} and \citet{mehri-etal-2019-structured} optimize the word-level policy by RL to improve task completion. Although these methods use RL to generate system utterances, they do not deal with the NLG module itself and ways to make it adaptive to environments and users.

\subsection{Adaptive Natural Language Generation for Task-oriented Dialogue}
Methods have been proposed for generating utterances adapted to the user. \citet{WALKER2004811} used quantitative user modeling for multimodal dialogue to achieve speech production that takes user preferences into account. \citet{janarthanam-lemon-2010-learning} used RL to create utterances that suit the user's domain knowledge. \citet{dusek-jurcicek-2016-context} proposed an NLG that can generate utterances exhibiting entrainment. Furthermore, \citet{mairesse2010towards} proposed PERSONAGE, an NLG that can generate utterances expressing Big Five personality traits. Our study differs from the above studies in that we optimize an existing NLG for the specific objective function of accurately conveying DAs for specific environments and users.

\begin{figure*}[t]
\centering
\includegraphics[scale=0.24]{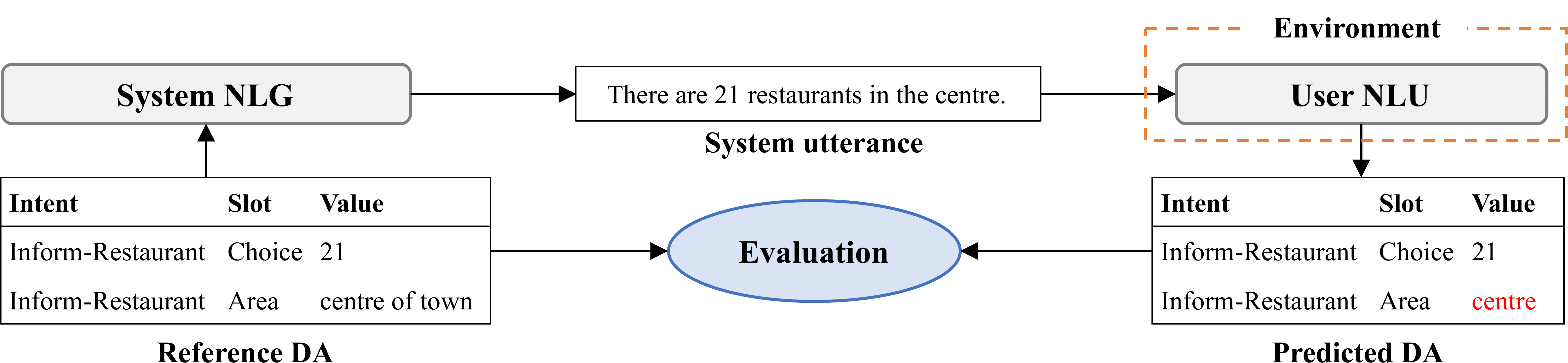}
\caption{Overview of the task to be performed by NLG. NLG generates system utterance corresponding to reference DA. NLG is evaluated using reference DA and predicted DA, which NLU estimates from system utterance. The ability of User NLU to understand can vary and so can the environment.}
\figlabel{task_overview}
\end{figure*}

\subsection{Natural Language Generation with Reinforcement Learning}
In recent years, many methods have been proposed that use RL for language generation tasks \citep{ijcai2019-880}. There are machine translation methods \citep{wu2016google, bahdanau2016actor} using BLEU as the reward function, summary generation methods \citep{ranzato2015sequence, dong-etal-2018-banditsum} using ROUGE, and story generation \citep{ijcai2019-829}. In addition, human feedback rather than automatic evaluation metrics is also used in many methods including machine translation \citep{kreutzer-etal-2018-neural}, summary generation \citep{ziegler2019fine, NEURIPS2020_1f89885d}, and open-domain dialogue \citep{hancock-etal-2019-learning, jaques2019way}. Our study examines the applicability of these recent advances to NLG in task-oriented dialogues.

\section{Method}
\subsection{Task Overview}
\seclabel{task_overview}
\figref{task_overview} shows the overall task performed by NLG in this study. First, NLG takes a reference DA, representing system intentions, converts it into natural language, and outputs a system utterance. The user's NLU then predicts the system's DA (predicted DA) from the system utterance. The goal of ANTOR is to generate utterances such that the predicted DA estimated by the NLU becomes equivalent to the reference DA. Note that this study uses automatic evaluation by comparing the reference DA and predicted DA; more down-to-earth evaluations using user subjective evaluations or user models are left for future work. In the following, the main concepts of the task, namely, DA, system NLG, user NLU, and evaluation, are described.

\paragraph{Dialogue Act} The DA is a semantic representation of a system utterance. The reference DA $A$ contains one or more triples consisting of intent $I$, slot $s$, and value $v$: 
$$A = \lbrace (I_1, s_1, v_1), ..., (I_{|A|}, s_{|A|}, v_{|A|}) \rbrace$$ 
$I$ represents a system's intention in a domain. For example, in the restaurant domain, there are intentions such as ``inform'' and ``request'' (e.g., ``Restaurant-Inform,'' ``Restaurant-Request''). $s$ and $v$ indicate the category (e.g., ``Choice'' and ``Area'') and specific information belonging to $s$, respectively. The first line of the reference DA in \figref{task_overview} indicates the semantics that there are 21 possible restaurants.

\paragraph{System NLG} NLG generates a system utterance $U$ on the basis of a given $A$. In this study, we assume a generative model with neural networks. Using the chain rule, a joint probability over $\lbrack A; U\rbrack = (x_1, ..., x_N)$ is modeled by a neural network $\rho_\theta$ with parameters $\theta$:
\begin{equation}
    \rho_\theta(\lbrack A; U \rbrack)=\prod^N_{n=1}\rho_\theta(x_n|x_{<n})
\end{equation}
where $N$ is the length of $\lbrack A; U\rbrack$. $\theta$ is trained by maximizing the log-likelihood (MLE) over dataset $D = \lbrace \lbrack A_1; U_1 \rbrack, ..., \lbrack A_{|D|}; U_{|D|} \rbrack \rbrace$:
\begin{equation}
\mathcal{L}(D) = \sum^{|D|}_{t=1}\sum^{N_t}_{n=1} \log \rho_\theta(x^t_n|x^t_{<n})
\equlabel{nlg_mle}
\end{equation}
where $N_t$ is the length of $\lbrack A_t; U_t \rbrack$.

\paragraph{User NLU} NLU predicts DA $A'$ from $U$ output by NLG. The structure of $A'$ is the same as that of $A$. In this study, we assume a classification-based prediction model for intent detection and slot tagging. Intent detection performs multi-label classification of an utterance, and slot tagging categorizes each token in an utterance as to which slot it belongs. The training data $D$ for NLG is the same as that for training NLU.

\paragraph{Evaluation} The goal of NLG is to generate $U$ such that $A = A'$. Therefore, the rate of concordance between $A$ and $A'$ is used to evaluate the NLG. Specifically, we use the F1 score calculated from true positive triples $A^{TP} = A \cap A'$, false negative triples $A^{FN} = A \cap \overline{A'}$, and false positive triples $A^{FP} = \overline{A} \cap A'$. In addition, following \cite{wang-etal-2020-data}, ${\rm Accuracy} = \frac{|A^{TP}|}{|A^{TP}|+|A^{FP}|+|A^{FN}|}$ is also used.

\subsection{Fine-tuning via Reinforcement Learning}
ANTOR is optimized by fine-tuning NLG pre-trained by MLE in \equref{nlg_mle} via RL. We use proximal policy optimization (PPO) \citep{schulman2017proximal} for the RL algorithm. We initialize policy $\pi_{\phi}$ by using $\rho_\theta$ and add a randomly initialized linear layer that outputs a scalar value for a value network. Parameters $\phi$ are updated on the basis of the clipped surrogate objective $\mathcal{L}^{CLIP}(\phi)$. We incorporate an understanding of NLU into the reward for ANTOR. When computing the reward $r$, each utterance $U$ generated by $\pi_{\phi}$ from $A \sim D$ is evaluated using $A'$ predicted by NLU from $U$ as follows:
\begin{equation}
    \footnotesize
    r(A, A')={\rm F1}(A, A') \frac{1}{|A^{TP}|}\sum _{(I,s,v)\in A^{TP}}{\rm idf}_{D}(I, s)
\end{equation}
where ${\rm idf}_{D}(I, s)$ is the IDF value of $(I, s)$ computed over all intent-slot pairs in $D$. This weighting of F1 scores compensates for DAs that frequently occur in $D$ (e.g., greetings) and DAs that occur infrequently.

Following \citet{ziegler2019fine}, to prevent $\pi_{\phi}$ from moving too far from $\rho_\theta$, a penalty by Kullback–Leibler (KL) divergence is added to $r(A, A')$ as the final reward $R$:
\begin{equation}
    R(A, A', U) = r(A, A') - \beta \log \frac{\pi_{\phi}(U|A)}{\rho_\theta(U|A)}
    \equlabel{compute_reward}
\end{equation}
where $\beta$ is the coefficient for the penalty. Algorithm \ref{alg:antor_with_ppo} summarizes the fine-tuning process of ANTOR. 

\begin{algorithm}
\caption{ANTOR with PPO}\label{alg:antor_with_ppo}
\begin{algorithmic}[1]
\parRequire{Dataset $D$; NLU; Policy $\rho_\theta$ pre-trained\\ via MLE by \equref{nlg_mle}}
\State Initialize policy $\pi_{\phi_{old}} = \rho_\theta$
\State Randomly initialize value network in $\pi_{\phi_{old}}$
\For{i = 1, 2, ..., max iteration}
\For{j = 1, 2, ..., batch size}
\State Sample a reference DA $A$ from $D$
\parState{Sample an utterance $U$ from $A$ by $\pi_{\phi_{old}}$}
\parState{Get a predicted DA $A'$ from $U$ by NLU}
\parState{Compute reward $R(A, A', U)$ by \equref{compute_reward}}
\State Compute advantage estimates
\EndFor
\parState{Optimize $\mathcal{L}^{CLIP}(\phi)$, with pre-determined number of epochs and minibatch size}
\State $\phi_{old} \leftarrow \phi$
\EndFor
\end{algorithmic}
\end{algorithm}

\section{Environment}
\seclabel{environmental_simulation}
We aim to confirm the feasibility of NLG that can robustly respond to the dialogue environment and the user, which does not exist in typical NLG training data. Therefore, we simulate two conditions in the following subsections. Note that $D$ in this section is the same as the data for training NLG in \secref{task_overview}; $D = \lbrace \lbrack U_1; A_1 \rbrack, ..., \lbrack U_{|D|}; A_{|D|} \rbrack \rbrace$.

\subsection{Speech Recognition Error}
\seclabel{speech_recognition_error}
\begin{figure}[t]
\centering
\includegraphics[scale=0.25]{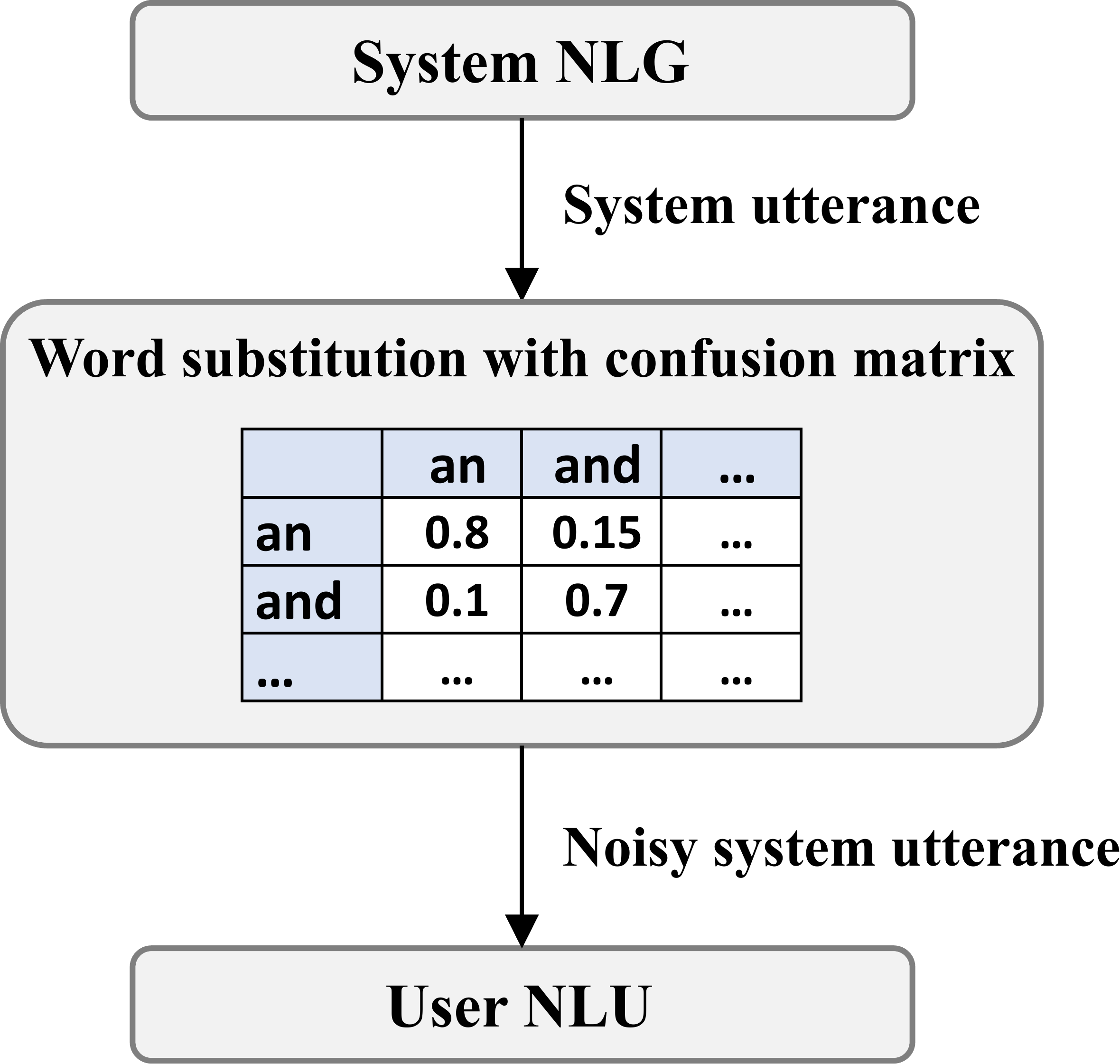}
\caption{ASR error simulation to add noise to system utterance}
\figlabel{asr_error_simulation}
\end{figure}
When a dialogue system interacts with a user via voice, it is assumed that background noise makes it difficult for system utterances to be accurately conveyed to the user. Automatic speech recognition (ASR) error simulation is often used to construct a noisy channel between a user and system \citep{schatzmann2007error, fazel2019investigation, wang-etal-2020-data}. Therefore, we apply perturbations that take the background noise into account to an utterance from NLG by using ASR error simulation. The noisy utterance is then used as input to NLU. Word substitution with a confusion matrix is used in the simulation (\figref{asr_error_simulation}). The TTS-ASR pipeline \citep{park19e_interspeech} is used to construct the confusion matrix with the following procedure:
\begin{enumerate}
\item Convert each $U$ over $D$ to audio data $U^{Audio}$ using a text-to-speech (TTS) system.
\item Create $U^{NoisyAudio}$ by adding background noise to each $U^{Audio}$.
\item Recover each $U^{NoisyAudio}$ into text $U^{Noisy}$ using an ASR system. This creates $D^{Noisy} = \lbrace U^{Noisy}_1, ..., U^{Noisy}_{|D|}\rbrace$.
\item Align words in $U$ and $U^{Noisy}$ with the Levenshtein distance and calculate the frequency of each word substituted and deleted from $U$ to $U^{Noisy}$, resulting in an $N$-dimensional confusion matrix $M\in N^2$. $N$ is the size of vocabulary $V = \lbrace v_1, ..., v_{N} \rbrace$ appearing in $D$ or $D^{Noisy}$. Note that a special token denoting deletion is included in $V$.
\end{enumerate}
Here, $M(i,j)$ indicates how often a word $v_i$ is substituted into $v_j$. When simulating ASR errors, each word $w$ in a system utterance is replaced by a word $v_j$ according to the following probability:
\begin{equation*}
  p_w(v_j)=
  \begin{cases}
    \frac{M(i,j)}{\sum^{N}_{n=1}M(i,n)} & \text{if $\exists v_i\in V: w=v_i$,} \\
    0 & \text{otherwise}
  \end{cases}
\end{equation*}

\subsection{Different Vocabulary Levels}
\seclabel{different_vocabulary_levels}
In a real environment in which a dialogue system interacts, the users may not have a sufficient vocabulary, such as when they are children or second language learners. Therefore, NLG should use vocabulary and sentences appropriate to the user's vocabulary level. We can simulate the user's vocabulary level by adjusting the training data $D$ for NLU as follows:
\begin{enumerate}
\item Prepare a word list $L=\lbrace v_1, ..., v_{|L|} \rbrace$ of the desired vocabulary level.
\item For each $\lbrack A; U \rbrack \in D$, if the lemma of a non-stop word\footnote{We used the Python library Spacy for word tokenization, stop word determination, and word lemmatization.} in $U$ is not in $L$, the $\lbrack A; U\rbrack$ is excluded from $D$.
\end{enumerate}
Using the adjusted training data, it is expected that the NLU can understand only the words in $L$.

\section{Experiments}
We wanted to confirm that ANTOR is capable of generating utterances adapted to the dialogue environment and the user. To verify the effectiveness of ANTOR, we conducted experiments using simulations.

\subsection{Dataset}
We used the MultiWOZ dataset \citep{budzianowski-etal-2018-multiwoz}, which is a task-oriented dialogue dataset between a clerk and a tourist at a tourist information center. The dataset contains 10,438
dialogues in seven domains. We used only system utterances annotated with the clerk's DAs. We used a total of 56,750 utterances and DA pairs in the training data of MultiWOZ to train NLG and NLU. In addition, we also used the utterances to construct the confusion matrix used in the ASR error simulation.

\subsection{Training Setup}
\paragraph{ANTOR} The 117M parameter version of the GPT-2 language model \citep{radford2019language} was used as a base model. DAs were input to the model as a sequence of intent, slot, and value triples connected by the symbols ``+'' and ``*''; if there were multiple triples, they were connected by commas ``, ''. In addition, to control generation, special tokens ``[ACT]'' and ``[RSP]'' were added at the beginning of the DAs and the system utterance sequences, respectively, in order to indicate the start of each sequence. The following is an example input to the model:
\begin{quote}
\small
\tt{[ACT] Inform-Restaurant + Choice * 21, Inform-Restaurant + Area * centre of town [RSP] There are 21 restaurants in the centre of town.}
\end{quote}

In MLE, GPT-2 was trained on MultiWOZ for five epochs with a batch size of 8, following \citet{peng2020few}. We used the Adam optimizer \citep{kingma2014adam} with a learning rate of 5e-5, and the learning rate decreased linearly with the number of steps.

For RL, 60 iterations were trained with a batch size of 1,024 (i.e., 1,024 utterances), and each batch was trained in 4 epochs with a minibatch size of 1. The coefficient $\beta$ of the penalty for the KL divergence was set to 0.1. We use generalized advantage estimation \citep{schulman2015high} (GAE) with a $\gamma$ of 1.0 and $\lambda$ of 0.95. The Adam optimizer was used with a learning rate of 5e-6, and the learning rate decreased linearly with the number of steps. 

For fair evaluation, we trained ANTOR with five different random seeds. 7,372 pairs of DAs and system utterances from MultiWOZ test data were used for testing. The average of the five trials was used as the final score. A greedy search was used for utterance generation in a test.

\paragraph{User NLU} As NLUs in our experiments, following the work of \citet{liu-etal-2021-robustness}, who evaluated NLUs with task-oriented dialogues, we used two models, MILU \citep{hakkani-tr2016multi-domain} and BERT \citep{devlin-etal-2019-bert}. Each model was trained by using pairs of DAs and system utterances from MultiWOZ training data. The learning rates were 1e-3 for MILU and 1e-4 for BERT as in \citep{liu-etal-2021-robustness}.

\subsection{Baselines}
To evaluate the performance of ANTOR, we used three comparison models.
\begin{description}
\item[SC-LSTM \citep{wen-etal-2015-semantically}] An LSTM-based method for controlling utterance generation with feature vectors related to DAs. We used a model pre-trained with MultiWOZ, which is available from ConvLab-2 \citep{zhu-etal-2020-convlab}, a platform for task-oriented dialogue systems.
\item[SC-GPT \citep{peng2020few}] A GPT-2 based model that has been trained on a large number of task-oriented dialogue datasets and further fine-tuned on MultiWOZ. In fine-tuning, training was done for five epochs with a batch size of 8, as reported in the official repository\footnote{\url{https://github.com/pengbaolin/SC-GPT}}.
\item[GPT-2 \citep{radford2019language}] A GPT-2 model fine-tuned on MultiWOZ using only MLE. The hyperparameters and input format were the same as those of ANTOR.
\end{description}

\subsection{Experimental Procedure}
The experiment was conducted in three stages using both MILU and BERT. First, we checked the effectiveness of ANTOR in clean environments with basic task-oriented dialogue. Next, we conducted two experiments: (1) in an ASR error simulation environment and (2) using NLUs trained only with low vocabulary levels.

\subsection{Results in Clean Environment}

\begin{figure}[t]
\includegraphics[scale=0.36]{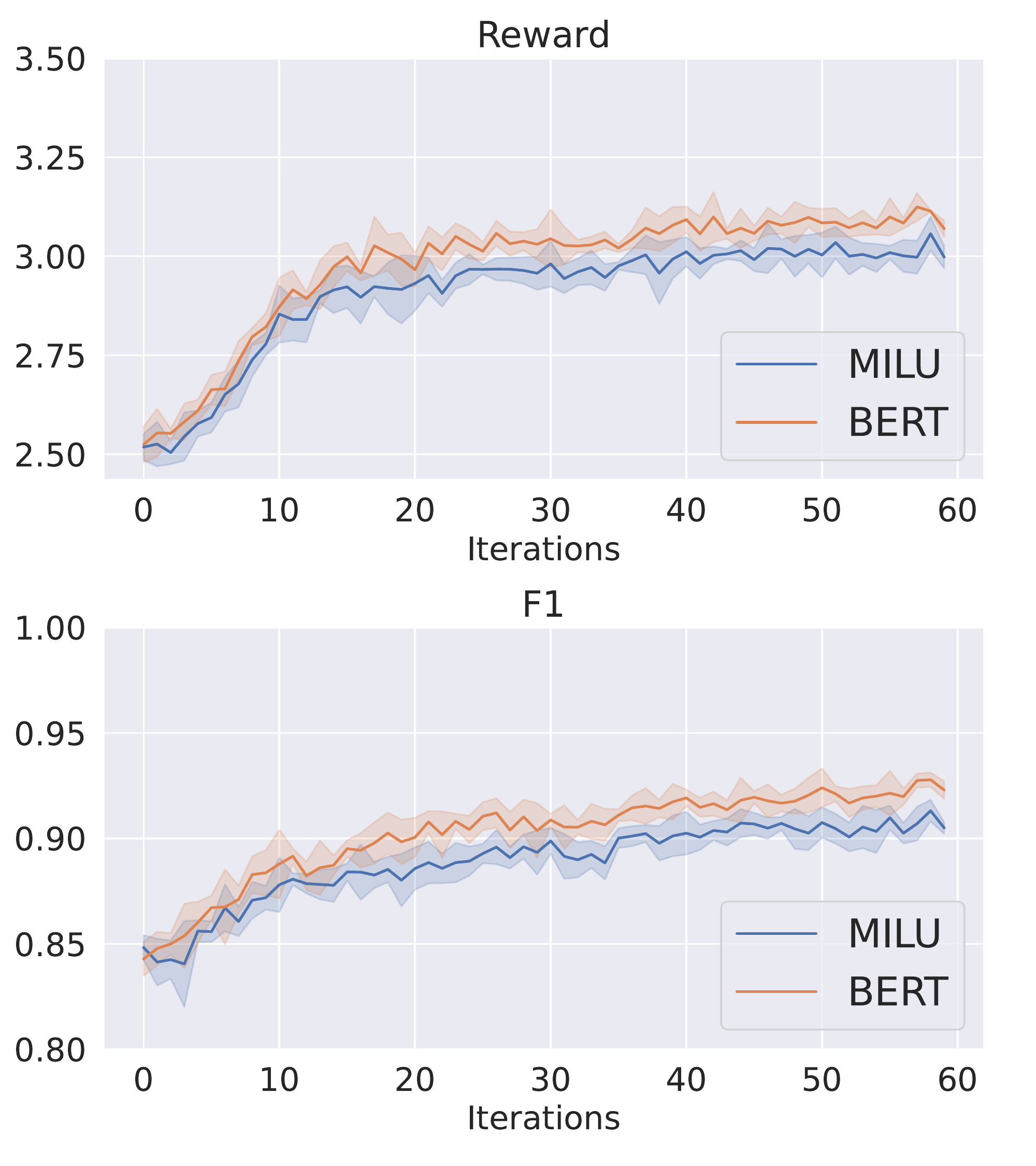}
\caption{Increase in reward and F1 when ANTOR was trained in a clean environment using MILU and BERT, respectively.}
\figlabel{clean_env_trans}
\end{figure}

\begin{table}[t]
\small
\centering
\begin{tabular}{lcc|ccc} \toprule
\mr{2}{\textbf{Model}} & \ml{2}{c|}{\textbf{MILU}} &\ml{3}{c}{\textbf{BERT}} \\ \cmidrule{2-6}
 & Acc. & F1 & Acc & F1 & BLEU \\ \midrule
SC-LSTM & 74.0 & 78.6 & 73.6 & 77.7 & 25.3 \\
SC-GPT & 77.3 & 81.1 & 78.3 & 82.2 & {\bf 29.9} \\ \midrule
GPT-2 & 79.5 & 84.0 & 79.7 & 83.6 & {\bf 29.9} \\
ANTOR (ours) & {\bf 86.7} & {\bf 89.8}& {\bf 87.8} & {\bf 90.7} & 27.5 \\ \bottomrule
\end{tabular}
\caption{Scores for each NLG model evaluated using MILU and BERT, respectively.}
\tbllabel{clean_env_result}
\end{table}

\figref{clean_env_trans} shows the reward and F1 transition of ANTOR, indicating that the scores increased steadily. \tblref{clean_env_result} shows the test scores for each model, indicating that ANTOR's accuracy and F1 outperformed the other models. These results show that ANTOR can learn utterance generation that fits both models of MILU and BERT. Note that the BLEU score of ANTOR was lower than those of SC-GPT and GPT-2. This BLEU score was calculated by comparing the utterances in MultiWOZ as references and the utterances generated by NLG as hypotheses. This means that ANTOR no longer generated utterances that appear in MultiWOZ in order to generate utterances tailored to NLU.

\subsection{Conditions for Speech Recognition Error}

\begin{table}[t]
\small
\centering
\begin{tabular}{ccccc} \\ \toprule
{\bf SNR} & {\bf WER} & {\bf Sub.} & {\bf Ins.} & {\bf Del.} \\ \midrule
0 & 30.4\% & 15.4\% & 8.5\% & 6.6\% \\
5 & 23.9\% & 12.6\% & 9.1\% & 2.2\% \\
10 & 21.5\% & 11.0\% & 9.2\% & 1.3\% \\
20 & 19.9\% & 9.8\% & 9.1\% & 1.0\% \\ \bottomrule
\end{tabular}
\caption{WER and percentage of error types for each SNR using TTS-ASR pipeline.}
\tbllabel{asr_error_env_wer}
\end{table}

\begin{table*}[t]
\small
\centering

\begin{subtable}{\textwidth}
\centering
\begin{tabular}{lccc|ccc|ccc|ccc} \toprule
\mr{3}{\textbf{Model}} & \ml{12}{c}{\textbf{SNR}} \\ \cmidrule{2-13}
 & \ml{3}{c|}{0} & \ml{3}{c|}{5} & \ml{3}{c|}{10} & \ml{3}{c}{20} \\ \cmidrule{2-13}
 & Acc. & F1 & WER & Acc. & F1 & WER & Acc. & F1 & WER & Acc. & F1 & WER \\ \midrule
SC-LSTM & 46.6 & 53.8 & 26.9 & 50.6 & 57.9 & 17.8 & 52.2 & 59.6 & 14.8 & 53.5 & 60.8 & 13.1 \\
SC-GPT & 47.9 & 55.4 & 27.2 & 52.1 & 59.9 & 18.0 & 54.3 & 61.9 & 14.9 & 55.1 & 62.8 & 13.3 \\ \midrule
GPT-2 & 48.2 & 56.0 & 28.2 & 52.8 & 60.6 & 19.2 & 54.9 & 62.7 & 16.0 & 56.0 & 63.7 & 14.3 \\
ANTOR (ours) & {\bf 51.9} & {\bf 59.4} & {\bf 26.8} & {\bf 56.9} & {\bf 64.2} & {\bf 18.5} & {\bf 59.9} & {\bf 66.9} & {\bf 14.8} & {\bf 60.9} & {\bf 67.9} & {\bf 13.7} \\ \bottomrule
\end{tabular}
\caption{MILU}
\end{subtable}

\hfil

\begin{subtable}{\textwidth}
\centering					
\begin{tabular}{lccc|ccc|ccc|ccc} \toprule
\mr{3}{\textbf{Model}} & \ml{12}{c}{\textbf{SNR}} \\ \cmidrule{2-13}
 & \ml{3}{c|}{0} & \ml{3}{c|}{5} & \ml{3}{c|}{10} & \ml{3}{c}{20} \\ \cmidrule{2-13}
 & Acc. & F1 & WER & Acc. & F1 & WER & Acc. & F1 & WER & Acc. & F1 & WER \\ \midrule
SC-LSTM & 46.6 & 53.2 & 27.1 & 50.3 & 57.4 & 18.0 & 52.4 & 59.5 & 14.9 & 53.4 & 60.4 & 13.0 \\
SC-GPT & 48.9 & 56.4 & 27.2 & 53.5 & 60.8 & 18.2 & 55.1 & 62.4 & 15.0 & 56.2 & 63.4 & 13.2 \\ \midrule
GPT-2 & 48.7 & 56.2 & 28.3 & 53.5 & 61.0 & 19.3 & 55.6 & 62.9 & 16.1 & 56.5 & 63.9 & 14.1 \\
ANTOR (ours) & {\bf 54.2} & {\bf 61.5} & {\bf 28.0} & {\bf 58.8} & {\bf 66.0} & {\bf 18.7} & {\bf 60.8} & {\bf 67.9} & {\bf 15.9} & {\bf 61.9} & {\bf 69.0} & {\bf 13.7} \\ \bottomrule
\end{tabular}
\caption{BERT}
\end{subtable}

\caption{Scores for methods evaluated in ASR error simulation environment with background noise at each SNR, using MILU and BERT, respectively. WER indicates how much error was imposed on NLG's output utterances.}
\tbllabel{asr_error_env_result}
\end{table*}

\begin{table*}[t]
\small
\centering
\begin{tabular}{llcc|cc|cc|cc} \toprule
\ml{2}{l}{\mr{2}{{\bf Model}}}& \ml{8}{c}{\textbf{CEFR-J level}} \\ \cmidrule{3-10}
 & & \ml{2}{c|}{$\leq$A1} & \ml{2}{c|}{$\leq$A2} & \ml{2}{c|}{$\leq$B1} & \ml{2}{c}{$\leq$B2} \\ \cmidrule{3-10}
NLG & NLU & Acc. & F1 & Acc. & F1 & Acc. & F1 & Acc. & F1 \\ \midrule
SC-LSTM & MILU & 47.4 & 53.9 & 55.7 & 62.5 & 62.4 & 68.8 & 63.6 & 69.8 \\
SC-GPT & MILU & 47.2 & 53.9 & 56.4 & 63.5 & 63.4 & 70.1 & 66.1 & 72.5 \\
GPT-2 & MILU & 48.4 & 55.0 & 57.7 & 64.7 & 66.0 & 72.6 & 68.6 & 74.8 \\
ANTOR (ours)& MILU & {\bf 54.7} & {\bf 61.1} & {\bf 63.5} & {\bf 69.8} & {\bf 72.5} & {\bf 78.0} & {\bf 75.0} & {\bf 80.1} \\ \midrule
SC-LSTM & BERT & 68.6 & 73.6 & 72.0 & 76.3 & 72.8 & 76.9 & 73.1 & 77.2 \\
SC-GPT & BERT & 70.3 & 75.9 & 73.3 & 77.9 & 77.3 & 81.4 & 77.6 & 81.9 \\
GPT-2 & BERT & 65.3 & 70.8 & 74.9 & 79.4 & 79.1 & 83.5 & 78.2 & 82.5 \\
ANTOR (ours)& BERT & {\bf 83.0} & {\bf 87.0} & {\bf 85.7} & {\bf 89.2} & {\bf 87.8} & {\bf 90.6} & {\bf 87.7} & {\bf 90.6} \\ \bottomrule
\end{tabular}
\caption{Scores for each NLG model when evaluated using MILU and BERT. Both MILU and BERT were trained using only vocabulary defined at each CEFR-J level.}
\tbllabel{lower_vocab_env_result}
\vspace{-2mm}
\end{table*}

\begin{table}[t]
\small
\centering
\begin{tabular}{lcccc} \\ \toprule
\mr{2}{\textbf{Model}} & \ml{4}{c}{\textbf{\% vocab. level in generation}} \\ \cmidrule{2-5}
 & $\leq$A1 & $\leq$A2 & $\leq$B1 & $\leq$B2 \\ \midrule
GPT-2 & 64.8 & 77.2 & 87.2 & 90.3 \\
ANTOR w/ MILU & {\bf 68.1} & {\bf 79.8} & {\bf 88.0} & {\bf 91.0}\\
ANTOR w/ BERT & 67.0 & 79.6 & {\bf 88.0} & {\bf 91.0}\\ \bottomrule
\end{tabular}
\caption{Percentage of vocabulary levels to which words generated by GPT-2 and ANTOR belong. ``w/ MILU'' and ``w/ BERT'' are ANTOR models trained with MILU and BERT, respectively. Note that ANTOR is fine-tuned using NLU trained on data at the CEFR-J level indicated by each column.}
\tbllabel{lower_vocab_env_level_percentage}
\vspace{-2mm}
\end{table}

We trained and evaluated ANTOR in an environment with ASR simulation. Google Cloud Text-to-Speech\footnote{\url{https://cloud.google.com/text-to-speech}} and Speech-to-Text\footnote{\url{https://cloud.google.com/speech-to-text}} were used for the TTS and ASR in the construction of the confusion matrix (see \secref{speech_recognition_error}). The ESC-50 dataset \citep{piczak2015dataset} was used as the background sound source, and it contains a total of 2,000 different sounds in five categories (e.g., natural soundscapes and urban noises). Randomly selected background noise was assigned to each utterance with a signal-to-noise ratio (SNR) of 0, 5, 10, and 20 dB. The range was selected so that the word error rate (WER) between original and noisy utterance text would be evenly distributed. \tblref{asr_error_env_wer} shows the WER of all of the data generated at each SNR and the percentages of substitution (Sub.), insertion (Ins.), and deletion (Del.) errors.

\tblref{asr_error_env_result} shows the evaluation results for each model. Overall, ANTOR showed a higher accuracy and F1 than all three comparison models. These indicate that ANTOR can preferentially generate words that are less likely to be confused. The above result shows that fine-tuning via RL enabled NLG to generate utterances adapted to the noisy environment, regardless of the noise intensity.

\subsection{Conditions for Different Vocabulary Levels}
We experimented with multiple NLUs trained on data that were gradually filtered along vocabulary levels. For word lists organized by vocabulary level, we used the Common European Framework of Reference (CEFR)'s English Vocabulary Profile\footnote{\url{https://www.englishprofile.org/}}. The CEFR defines six levels of language acquisition, from A1 (beginner) to C2 (proficient, comparable to native speakers), with a word list for each level. In our experiment, we used the CEFR-J \citep{tono2012cefr} word list for Japanese-English learners\footnote{\url{http://www.cefr-j.org/download_eng.html}}. We created four types of training data for NLU by filtering MultiWOZ data with a focus on the four levels $\leq$A1, $\leq$A2, $\leq$B1, and $\leq$B2 (see \secref{different_vocabulary_levels}). Note that C1 and C2 were not available in CEFR-J and were not used in our experiment. As a result of the filtering, the number of utterances in the datasets at the $\leq$A1, $\leq$A2, $\leq$B1, and $\leq$B2 levels was 11,190, 15,538, 24,311, and 28,999 utterances, respectively. 

ANTOR was trained and evaluated using MILU and BERT trained on each of the four vocabulary levels. \tblref{lower_vocab_env_result} shows the results. ANTOR outperformed all three comparison models at all vocabulary levels, both with MILU and with BERT. In particular, when using BERT at level $\leq$A1, the original GPT-2's accuracy and F1 were lower than SC-LSTM and SC-GPT. However, ANTOR had an accuracy and F1 that were significantly improved over the GPT-2 scores by 17.7\% and 16.2\%, respectively, and it outperformed the scores of SC-LSTM and SC-GPT.

We checked whether the vocabulary in the utterances generated by ANTOR actually changed due to RL. \tblref{lower_vocab_env_level_percentage} shows the percentage of the vocabulary in the utterances that ANTOR had generated during the evaluation when trained with each NLU of each level. Note that stop words and proper nouns were excluded from the calculation. We see that both ANTOR w/ MILU and w/ BERT generated words at each level with a higher frequency than the original GPT-2. From these results, it is considered that the NLG was able to learn utterance generation tailored to the NLU's ability to understand.

\section{Case Study}
\begin{table}[t]
\small
\centering
{\tabcolsep=1.2mm
\begin{tabular}{lcrcr} \toprule
\textbf{Intent-slot pair} & \textbf{Num.} & \ml{3}{l}{\textbf{\% TP change}} \\ \midrule
(Request-Taxi, Depart) & 142 & 14.5 & $\rightarrow$ & 90.4 \\
(OfferBook-Train, People) & 6 & 0.0 & $\rightarrow$ & 66.7 \\
(Recommend-Hotel, Postcode) & 12 & 14.3 & $\rightarrow$ & 80.0\\
(Recommend-Restaurant, Price) & 49 & 20.8 & $\rightarrow$ & 84.0\\
(NoOffer-Hotel, none) & 32 & 23.5 & $\rightarrow$ & 86.7\\ \bottomrule
\end{tabular}
}
\caption{Top five intent-slot pairs that were correctly recognized by BERT at a higher percentage (\% TP) by ANTOR compared with GPT-2. ``Num.'' indicates the number of times each intent-slot pair appeared during test.}
\tbllabel{clean_env_tp_change}
\vspace{-2mm}
\end{table}

To see how ANTOR improved the performance of NLU, we analyzed the behavior of ANTOR. In this analysis, we used BERT because we thought that a difference from GPT-2 could be clearly seen since the F1 was improved more by using BERT than MILU as in \tblref{clean_env_result}. The case studies here are done in clean environments, but similar behaviors were observed for noisy environments and different user vocabulary levels.

\begin{table}[t]
\small
\centering

\begin{subtable}{0.48\textwidth}
\centering
\includegraphics[scale=0.26]{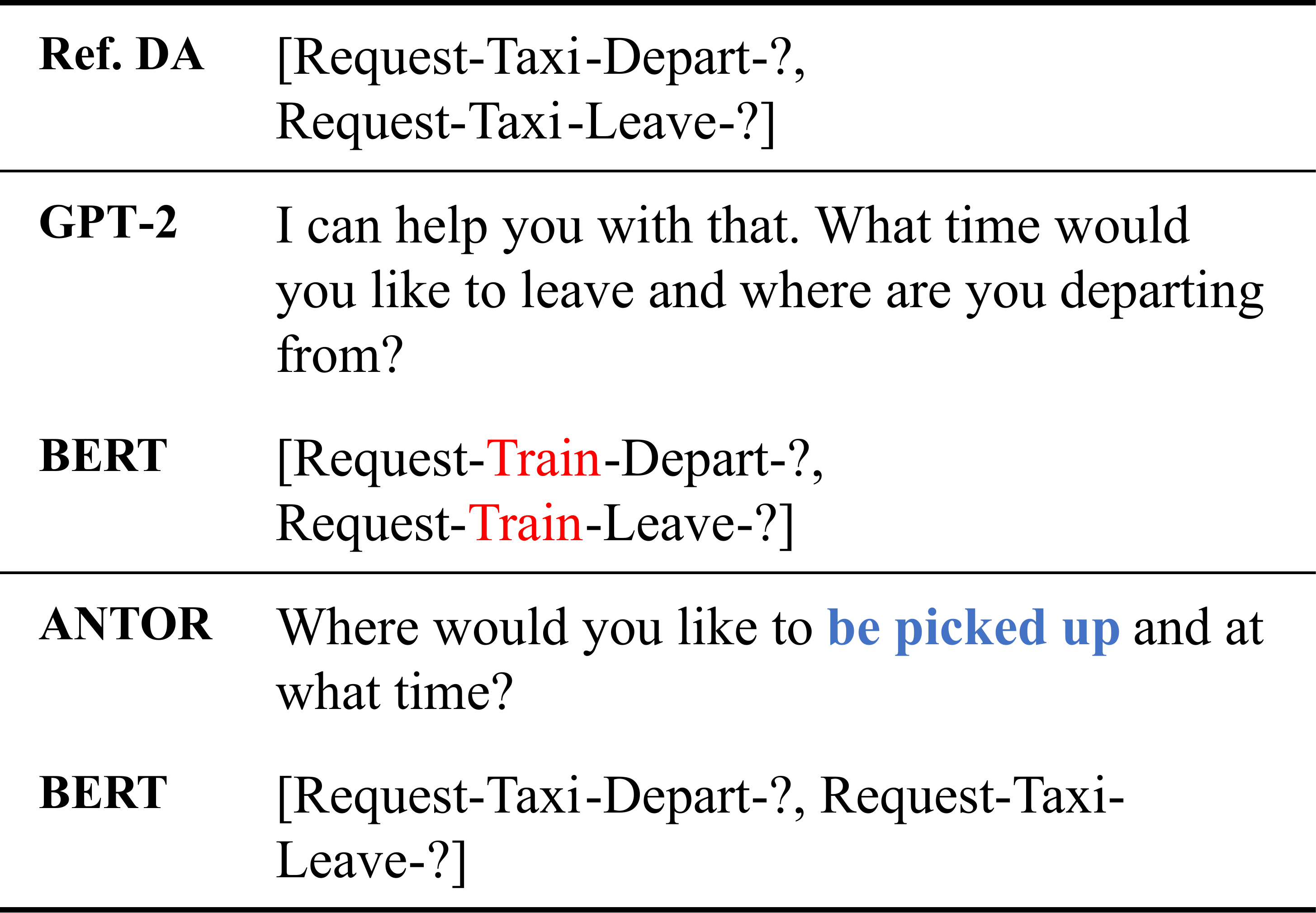}
\caption{Case 1}
\end{subtable}

\hfil

\begin{subtable}{0.48\textwidth}
\centering					
\includegraphics[scale=0.26]{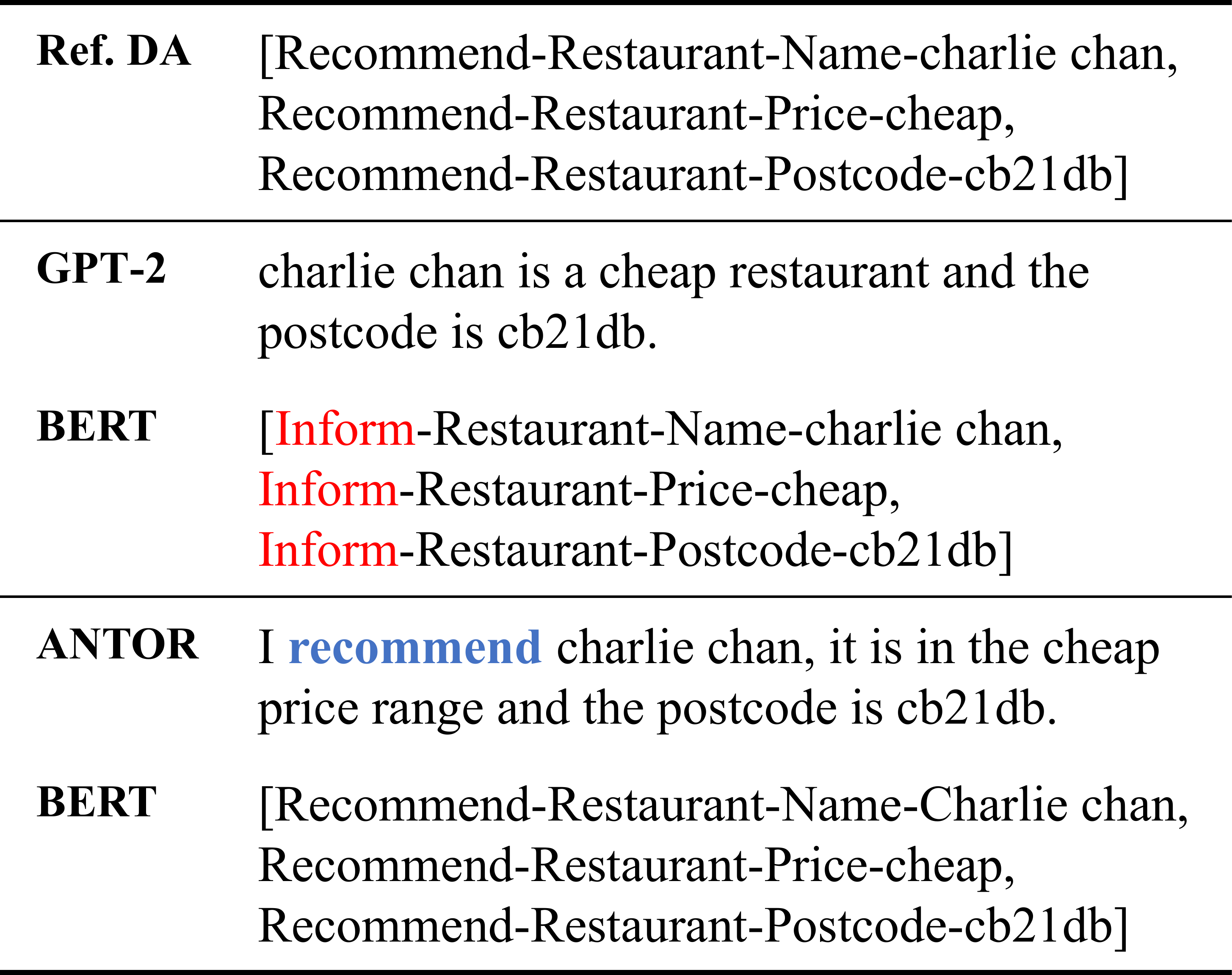}
\caption{Case 2}
\end{subtable}

\caption{Examples of utterances generated by GPT2 and ANTOR from DAs and DAs predicted by BERT for each utterance. Letters in red indicate DAs misrecognized by BERT. Letters in blue indicate words that may have influenced BERT's prediction.}
\tbllabel{clean_env_case_study}
\vspace{-2mm}
\end{table}

First, we listed the intents and slots in the DAs for which the NLU prediction accuracy was considerably improved by the utterances of ANTOR compared with those of GPT-2 (\tblref{clean_env_tp_change}). Next, we examined the utterances that each model generated from the listed DAs. \tblref{clean_env_case_study} shows examples of utterances generated for (Request-Taxi, Depart) and (Recommend-Restaurant, Price), which have a particularly high occurrence as in \tblref{clean_env_tp_change}. In case 1, BERT misidentified the train domain instead of the taxi domain from the GPT-2 utterances. In contrast, ANTOR correctly conveyed the DAs by explicitly using the phrase ``be picked up.'' In case 2, the intention of ``inform'' was conveyed by GPT-2 instead of ``recommend.'' However, ANTOR explicitly used the word ``recommend'' to correctly convey the DA. These results suggest that fine-tuning NLG using RL enables NLG to generate utterances adapted to the NLU.

Note that since humans will not have the problems that the NLU had here because humans have a better understanding, we expect ANTOR to adapt differently when interacting with humans.

\section{Summary and Future Work}
This paper investigated whether NLG can generate utterances adapted to the dialogue environment and the user via RL. We proposed a method, ANTOR, and conducted experiments using MultiWOZ to confirm that ANTOR can generate such utterances for multiple NLUs with different model architectures. In addition, we also consistently confirmed the effectiveness of ANTOR for noisy environments and a user's vocabulary levels.

For future work, we plan to evaluate whether ANTOR optimized for NLU is also effective for humans. We are also interested in extending our method for practical use (e.g., real-time adaptation to users in an online dialogue environment). Furthermore, we would like to utilize methods to optimize an entire system with RL, such as \cite{mehri-etal-2019-structured} and \cite{ohashi-2022-post}, so that all modules of a system can be adapted to users.

\section*{Acknowledgments}
This work was supported by JSPS KAKENHI Grant Number 19H05692. We used the computational resources of the supercomputer ``Flow'' at the Information Technology Center, Nagoya University.

\bibliography{custom}
\bibliographystyle{acl_natbib}

\end{document}